\documentclass[sigconf]{acmart}

\copyrightyear{2024}
\acmYear{2024}
\setcopyright{rightsretained}
\acmConference[KDD '24]{Proceedings of the 30th ACM SIGKDD Conference on Knowledge Discovery and Data Mining}{August 25--29, 2024}{Barcelona, Spain}
\acmBooktitle{Proceedings of the 30th ACM SIGKDD Conference on Knowledge Discovery and Data Mining (KDD '24), August 25--29, 2024, Barcelona, Spain}\acmDOI{10.1145/3637528.3671611}
\acmISBN{979-8-4007-0490-1/24/08}

\usepackage{booktabs} 
\usepackage{hyperref}
\usepackage{hyperxmp}
\newtheorem{definition}{Definition} 
\usepackage{subcaption}
\usepackage[bottom]{footmisc}
\usepackage{makecell}
\usepackage{setspace}
\usepackage{multirow}
\usepackage{threeparttable}
\usepackage{amsthm,amsmath}
\usepackage{color}
\usepackage{mathrsfs}
\usepackage{bm}

\begin{document}

\title{PAIL: Performance based Adversarial Imitation Learning Engine for Carbon Neutral Optimization}
\thanks{$^{+}$This work was accomplished when the first author working as intern in NEC Labs America supervised by the second author.}
\thanks{* Corresponding authors.}
\renewcommand{\shorttitle}{PAIL for Carbon Neutral Optimization}

\author{Yuyang Ye}
\affiliation{%
  \institution{Department of Management Science and Information Systems, Rutgers Business School, Rutgers University}
  \city{Newark}
  \state{NJ}
  \country{USA}}
\email{yuyang.ye@rutgers.edu}

\author{Lu-An Tang}
\authornotemark[1]
\affiliation{%
  \institution{Department of Data Science and System Security, NEC Laboratories}
  \city{Princeton}
    \state{NJ}
  \country{USA}}
\email{ltang@nec-labs.com}

\author{Haoyu Wang }
\affiliation{%
  \institution{Department of Data Science and System Security, NEC Laboratories}
  \city{Princeton}
    \state{NJ}
  \country{USA}}
\email{haoyu@nec-labs.com}

\author{Runlong Yu}
\affiliation{%
  \institution{Department of Computer Science, University of Pittsburgh}
  \city{Pittsburgh}
  \state{PA}
  \country{USA}}
\email{ruy59@pitt.edu}

\author{Wenchao Yu}
\affiliation{%
  \institution{Department of Data Science and System Security, NEC Laboratories}
  \city{Princeton}
    \state{NJ}
  \country{USA}}
\email{wyu@nec-labs.com}

\author{Erhu He}
\affiliation{%
  \institution{Department of Data Science and System Security, NEC Laboratories}
  \city{Princeton}
    \state{NJ}
  \country{USA}}
\email{ehe@nec-labs.com}

\author{Haifeng Chen}
\authornotemark[1]
\affiliation{%
  \institution{Department of Data Science and System Security, NEC Laboratories}
  \city{Princeton}
    \state{NJ}
  \country{USA}}
\email{haifeng@nec-labs.com}

\author{Hui Xiong}
\affiliation{%
  \institution{Thrust of Artificial Intelligence, The Hong Kong University of Science and Technology (Guangzhou)\\
Department of Computer Science and Engineering, The Hong Kong University of Science and Technology}
\city{Guangzhou}
\country{China}}
\email{xionghui@ust.hk}

\renewcommand{\shortauthors}{Yuyang Ye et al.}

\thispagestyle{plain}

\begin{abstract}
Achieving carbon neutrality within industrial operations has become increasingly imperative for sustainable development. It is both a significant challenge and a key opportunity for operational optimization in industry 4.0. In recent years, Deep Reinforcement Learning (DRL) based methods offer promising enhancements for sequential optimization processes and can be used for reducing carbon emissions. However, existing DRL methods need a pre-defined reward function to assess the impact of each action on the final sustainable development goals (SDG). In many real applications, such a reward function cannot be given in advance. To address the problem, this study proposes a Performance based Adversarial Imitation Learning (PAIL) engine. It is a novel method to acquire optimal operational policies for carbon neutrality without any pre-defined action rewards.  Specifically, PAIL employs a Transformer-based policy generator to encode historical information and predict following actions within a multi-dimensional space. The entire action sequence will be iteratively updated by an environmental simulator. Then PAIL uses a discriminator to minimize the discrepancy between generated sequences and real-world samples of high SDG. In parallel, a Q-learning framework based performance estimator is designed to estimate the impact of each action on SDG. Based on these estimations, PAIL refines generated policies with the rewards from both discriminator and performance estimator. PAIL is evaluated on multiple real-world application cases and datasets. The experiment results demonstrate the effectiveness of PAIL comparing to other state-of-the-art baselines. In addition, PAIL offers meaningful interpretability for the optimization in carbon neutrality.
\end{abstract}
\vspace{-4mm}



\maketitle


%

\section{Introduction}

In industry 4.0 era, the imperative of achieving carbon neutrality is closely intertwined with the pursuit of economic efficiency~\cite{ericsson2022sustainability}. The evolving carbon credit systems, designed to regulate and monetize carbon emissions, have constituted a significant component of overall economic measurement~\cite{gupta2011carbon}. The operational optimization tasks of industry 4.0 are required to achieve the dual objectives of economic efficiency and environmental sustainability~\cite{lasi2014industry}. This shift towards sustainable practices emphasizes not only the importance of advanced automation but also the technical innovation for carbon friendly operations. For example, many industry plants have deployed numerous sensors to collect environmental data like temperature and pressure. Based on the streaming data, the existing operating systems can automatically conduct actions to maximize the production efficiency. Such systems are required to balance the goals of product maximization and carbon emission minimization for sustainable development goals ~\cite{li2015energy}. Therefore, designing and developing a new operational system capable of making effective decisions to balance both economic and environmental objectives and achieve the sustainable development goals (SDG), have emerged as a major concern across many industrial applications.



Unfortunately, to the best of our knowledge, there is no solution for SDG optimization in industry systems yet, partially due to the following challenges. 
\vspace{-5.5mm}
\begin{itemize}
    \item \textbf{Historical Dependency}: The challenge of optimizing Sustainable Development Goals (SDG) in industrial systems is markedly accentuated by the historical dependency of action rewards. In these settings, the impact of actions are deeply intertwined with past operations, complicating the reward estimation process. Traditional Deep Reinforcement Learning (DRL)~\cite{liu2020parallel, han2020enabling} methods, which rely on predefined Key Performance Indicators (KPIs) for rewards, struggle in the context of carbon emissions related SDGs. The SDG objectives, unlike straightforward economic metrics, cannot be easily quantified on a step-by-step basis but are rather assessed cumulatively after the whole operational sequences. This historical interdependency complicates the identification of individual action rewards, challenging accurate reward estimation and optimization efforts.
    \item \textbf{Complex Action Space}: The optimization of SDGs involves navigating a complex, multi-dimensional continuous action space, characterized by the combination of various action types and operational values. This complexity is heightened by the dynamic nature of the impacts each action may have, further influenced by preceding actions within this continuous space. Accurately estimating the rewards for actions in such a nuanced and variable environment poses a significant challenge, necessitating sophisticated optimization algorithms that can adeptly manage the intricacies of action spaces in the industrial systems.
    \item \textbf{Sequence Diversity}: In SDG optimization within multi-dimensional continuous action spaces, rapid development of imitation learning (IL)~\cite{ho2016generative} allows for learning from high-performance policies without predefined reward functions. However, this approach struggles with sequence diversity and the vague definition of high performance in industrial settings. The scarcity of high-performance examples and the inherent variability in operational sequences challenge the generalization of models learned through IL. Though these IL based methods could avoid the complexity of designing reward functions, they may fail to adapt or generalize across varied industrial environments due to the diversity of actions and ambiguity in identifying optimal strategies.
\end{itemize}
To address the aforementioned challenges for carbon neutrality in industrial systems, this study proposes a Performance based Adversarial Imitation Learning (PAIL) framework. PAIL uses a sliding window Transformer framework to address the challenge of historical dependency. This framework is precisely designed to learn policies for predicting current action based on previous operations and the current state of the system. More specifically, to tackle the intricate action space, our approach derive action arrays, where each element denotes an action type, by probabilistic sampling from the multiple dimensional Gaussian Mixed Distributions produced by the encoder. Consequently, the future state would simultaneously be simulated with a pre-defined autoencoder. The entire action sequence can be iteratively updated in this way. Following designing an correspondingly adpated discriminator, we introduce an innovation to combat the sequence variability and generalization issues, which is archived by a Q-learning based performance estimator model. This performance estimator is tasked with instantaneously estimating the value of each action in terms of its contribution to the overall SDG, thereby enhancing the carbon neutral ability of distinct actions. Accordingly, the reward signal from both the discriminator and performance estimator collectively refine the generated policy. We quantitatively validate the effectiveness of PAIL using two real-world dataset, demonstrating that PAIL improves the overall performance of the industrial system and highlights the significance of key actions.

The technique contributions of this study are listed as follows:
\begin{itemize}
    \item We address carbon neutral optimization by developing a system for adjusting industrial processes, innovatively moving beyond traditional methods that depend on predefined reward functions. This approach offers a novel, data-driven solution to the challenge.
    \item We introduce the Performance-based Adversarial Imitation Learning Engine (PAIL), featuring a Transformer-based policy generator for crafting actions from historical data while refining the policy with incorporating an adapted discriminator and a novel Q-learning based performance estimator.
    \item We conducted extensive experiments on two real-world datasets for performance evaluation. The experimental results clearly validate the effectiveness of our approach compared to the state-of-the-art baselines. Furthermore, a detailed case study on industrial process analysis further clarifies the practical application and potential of our method.
\end{itemize}
The rest of this paper is organized as follows. Section 2 introduces the preliminaries of our problem. Section 3 describes our proposed PAIL solution. Section 4 empirically evaluates our model on real-world data. We summarize the related work in Section 5, followed by the conclusions in Section 6.
\section{Preliminary}
\begin{table}[b]
\caption{Notations}
\begin{tabular}{|l|l|}
\hline
\textbf{Notation}                              & \textbf{Descriptions}                                                                                                                                                                                                                    \\ \hline

$\mathcal{S}=\{s_t\}$                          & The set of environment or system state.                                                                                                                                                                                                   \\ \hline
$\mathcal{A}=\{a_t\}$                          & \begin{tabular}[c]{@{}l@{}}The actions performed on system, $\mathcal{A} \in \{R^+\}^K$\\ where $k^{th}$ element denotes the $k^{th}$ type of \\ operation and its value indicates the value of \\ corresponding operation.\end{tabular} \\ \hline
$\mathcal{H}=\{h_t\}$                          & \begin{tabular}[c]{@{}l@{}}The historic state and action pair at step $t$, wh-\\ere $h_t=(s_{t-l}, a_{t-l}, \cdots,  s_{t-1}, a_{t-1})$, $l$ is lookback.\end{tabular}                                                                                                        \\ \hline
$\mathrm{T}=\{\tau\}_n$                        & \begin{tabular}[c]{@{}l@{}}The set of trajectories consisting of the sequen-\\ces of states and actions where $\tau=(s_1, a_1, \cdots)$.\end{tabular}                                                                                      \\ \hline
$Y = {y}_n$                                    & The final SGD values for each trajectory.                                                                                                                                                                                                 \\ \hline
$\pi_E (a_t|s_t, h_t)$                     & The expert policy with high SGD.                                                                                                                                                                                                                        \\ \hline
$\pi_{\theta} (a_t|s_t, h_t)$              & The learnt policy parameterized by $\theta$.                                                                                                                                                                                              \\ \hline
$\rho_{\pi}:  \mathcal{S} \times  \mathcal{A}$ & \begin{tabular}[c]{@{}l@{}}The distribution of state-action pairs that the\\ policy $\pi$ interacts with the system.\end{tabular}                                                                                                         \\ \hline
\end{tabular}
\label{not}
\end{table}
\subsection{Problem Definition}
In industrial systems, measurements that are instantly readable can serve as states, alongside actionable adjustments. These states and actions enable us to optimize towards Sustainable Development Goals (SDGs), focusing on achieving carbon neutrality. Our objective is to enhance SDG through strategic actions on the system. Here, we first list the notations related to our problem as Table~\ref{not}. 

\begin{figure*}[t]
    \centering
    \includegraphics[width=0.85\linewidth]{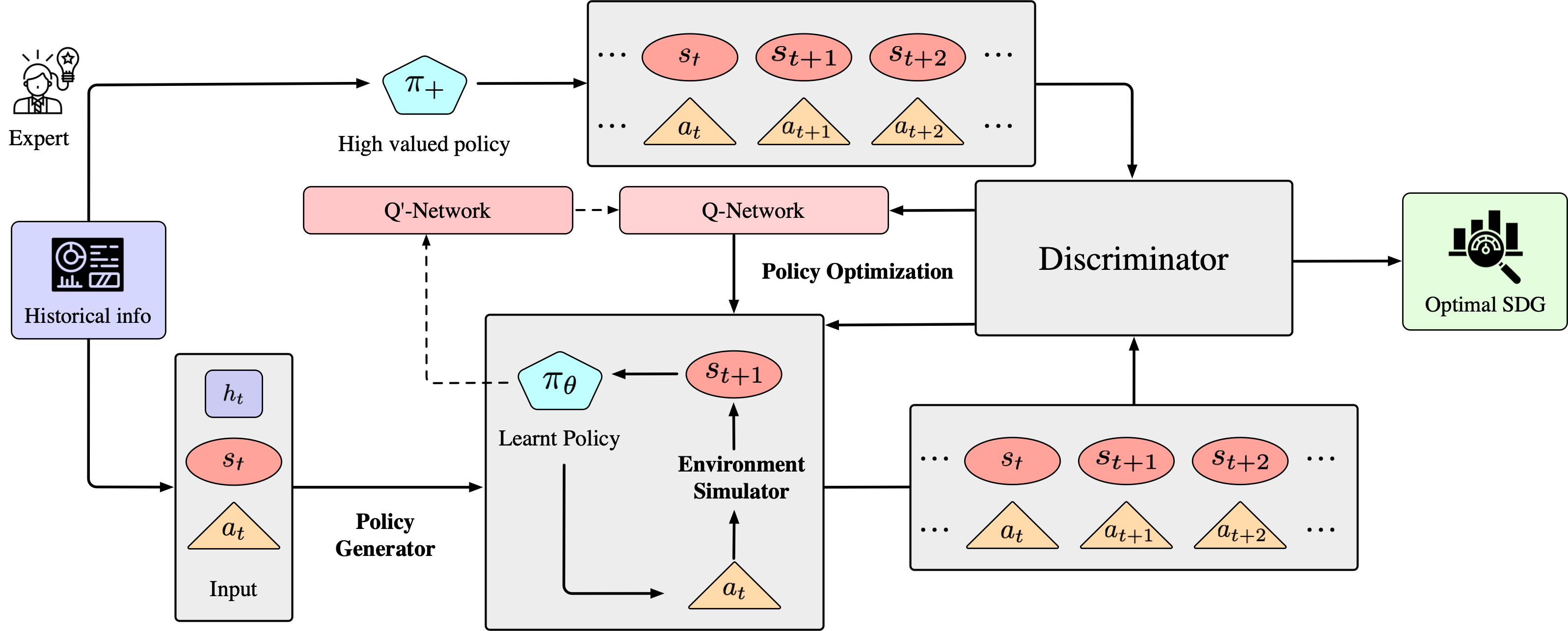}
    \caption{The PAIL framework for carbon neutral optimization.}
    \label{PAIL}
\end{figure*}

The primary objective of proposed solution is to optimize the actions performed in an industrial system and achieve the highest SDG. During the model learning phase, we should estimate the improvement of SDG by taking the learned action as a reward signal. On the other hand, we use the improvement of SDG as a performance indicator during the evaluation phase. To facilitate this process, we also need to learn a function capable of predicting the SGD value over time. Consequently, we divide the task of carbon neutral optimization into two sub-problems as follows.
\begin{definition}[Problem Definition]
Given a set of industrial operation trajectories denoted by $T = \{\tau_1, \tau_2, \ldots, \tau_n\}$, where each trajectory $\tau_i$ is a sequence of actions and states in the industry system, associated with a corresponding Sustainable Development Goal (SDG) $y_i$. The objective is twofold: (1) To predict the final SDG for a given complete trajectory, represented as $\hat{y} = f(\tau)$. (2) Let $t_k$ be the timestamp of starting inference, for each $t>t_k$, to learn an optimal policy $\pi_{\theta}(a_t \mid s_t, h_t)$ that infers action $a_t$ based on the current state $s_t$ and historical context $h_t$, aiming to reach the highest SDG.
\end{definition}

Without loss of generality, we assumes all trajectories have a uniform length $T$. In the pre-processing step, the system can align the trajectories of different lengths by adding dummy time windows in the end. In real applications, we usually leave the first 20-30\% time as "only monitoring". After the timestamp of "starting inference", the system uses all the historical data and current state to generate the action to be carried out in current timestamp. By iteratively obtaining the successor state from probability distribution, the system can produce an action sequence for all the remaining time. 

\subsection{Imitation Learning}
Generative Adversarial Imitation Learning (GAIL) is a method that integrates both behavior cloning and inverse reinforcement learning using a Generative Adversarial Networks (GAN)~\cite{goodfellow2020generative} based structure, which aims to minimize the Jensen-Shannon divergence between the expert policy $\pi_E$ and the generated policy $\pi_{\theta}$.

\begin{equation}
D_{JS}(\pi_E || \pi_{\theta}) = \frac{1}{2} D_{KL}(\pi_E || \frac{\pi_E + \pi_{\theta}}{2}) + \frac{1}{2} D_{KL}(\pi_{\theta} || \frac{\pi_E + \pi_{\theta}}{2})
\end{equation}

This model consists of two key components: a discriminator $D$ and a policy generator $\pi_{\theta}$. The generator aims to generate actions that mimic the expert behavior, while the discriminator aims to distinguish between the agent actions and the expert actions. Formally, the discriminator seeks to solve the optimization problem as follows,

\begin{equation}
\max_{D \in (0,1)^{\mathcal{S} \times \mathcal{A}}} \mathbb{E}_{\pi_{E}}[\log D(s, a)] + \mathbb{E}_{\pi_{\theta}}[\log(1-D(s, a))],
\end{equation}
where $\pi_{E}$ is the expert's policy, $\pi$ is the policy of the agent, and $(s, a)$ are state-action pairs. In contrast, the generator aims to fool the discriminator by producing actions that are indistinguishable from those of the expert, achieved by solving the following problem,

\begin{equation}
\min_{\pi_{\theta}} \mathbb{E}_{\pi}[\log(1-D(s, a))] + \lambda H(\pi),
\end{equation}
where $H(\pi)$ is the entropy of the policy, and $\lambda \geq 0$ is a coefficient that encourages exploration by the agent. By alternating between training the discriminator and the generator, GAIL learns a policy $\pi$ that is able to mimic the expert's behavior. Following the idea of imitation learning (IL), this work delves into the industrial context of carbon neutrality, proposing a novel algorithm designed to tackle the specific challenges previously outlined.

\section{Methodology}
In this section, we provide the technical details of proposed PAIL solution. As shown in Figure~\ref{PAIL}, PAIL has three major modules, namely Policy Generator, Environment Simulator, and Policy Optimization.

\subsection{Policy Generator}
\begin{figure}[t]
    \centering
    \includegraphics[width=0.7\columnwidth]{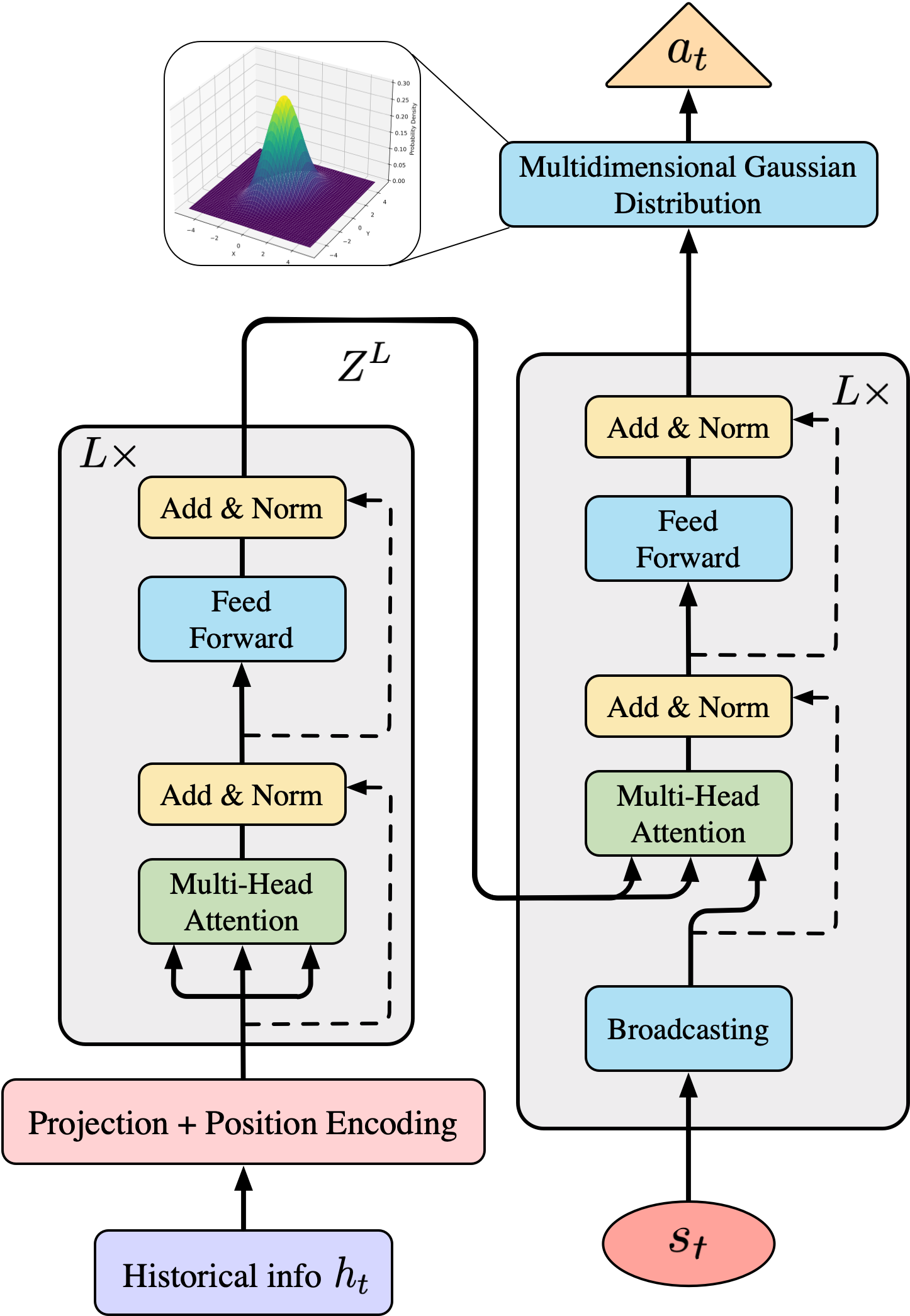}
    \caption{The framework of policy generator.}
    \label{policy}
\end{figure}

In many industrial contexts, the decision-making process is significantly influenced by historical states and actions. Take oil extraction as an example, the prior action, such as shutting off a gas valve, can significantly shape the probabilities of subsequent actions on the same equipment (e.g., shutting it off again or initiating a lift). In light of such critically important temporal dependencies, we design an adapted Transformer architecture to generate the action sequences with discerning temporal correlations in the trajectory, as shown in Figure~\ref{policy}. The multi-head self-attention architecture enables simultaneous processing of multiple trajectories. It is particularly beneficial for the capture of subtle and long-range inter-dependencies in the trajectory. This attribute is essential for precise modeling along temporal dimension that context and historical trends are paramount. Furthermore, given the task of forecasting new action sequences with historic trajectories, the self-attention mechanism exhibits superiority to dynamical adjusting focused segments of the input. 

For each trajectory $\tau_i$, let us partition it by a fixed window length $T$. And we get a window sequence \( \mathbf{X}_i = \{x_1, x_2, \dots, x_T\} \). Here each $\mathbf{x}_t \in \mathbf{X}_i$ includes two factors: the concatenated state $\mathbf{s}_t$ and action vector $\mathbf{a}_t$ in the window. Note that, the length of the input sequence $X_t$, representing historical information, varies for each time step $t$. To address this challenge, a sliding window methodology is employed to select the preceding $l$ elements for action prediction, here $l$ is a hyper-parameter. Thus, the historical information at time step $t$ can be written as $h_t=\{x_{t-l},\cdots,x_{t-1}\}$.

The next step is to project the input $\mathbf{H}_i = \{h_1, \cdots, h_T\}$ into spaces of query $\mathbf{Q}$, key $\mathbf{K}$, and value $\mathbf{V}$ via projection matrices \( \mathbf{W}_q \), \( \mathbf{W}_k \), and \( \mathbf{W}_v \) $\in \mathbb{R}^{d \times d}$. The correlation across time steps within the sequence is computed as follows.
\begin{equation}
\mathbf{Z}_i = \text{softmax}\left(\frac{\mathbf{Q}\mathbf{K}^T}{\sqrt{d_k}}\right)\mathbf{V} = \text{softmax}\left(\frac{(\mathbf{H}_i\mathbf{W}_q)(\mathbf{H}_i\mathbf{W}_k)^T}{\sqrt{d_k}}\right)(\mathbf{H}_i\mathbf{W}_v).
\end{equation}

In the foundational Transformer model, positional information is incorporated using sinusoidal position encoding. In the Transformer framework, these representations traverse $L$ layers, and are subjected to both multi-head self-attention and position-wise feed-forward operations. Formally, for each layer \( l = 1, \dots, L \):

\begin{align}
\mathbf{Z}^l &= \text{LayerNorm}\left(\mathbf{Z}^{l-1} + \text{MultiHead}(\mathbf{Q}^{l-1}, \mathbf{K}^{l-1}, \mathbf{V}^{l-1})\right), \\
\mathbf{Z}^l &= \text{LayerNorm}\left(\mathbf{Z}^l + \text{FFN}(\mathbf{Z}^l)\right).
\end{align}

The multi-head attention mechanism partitions $\mathbf{Z}$ into $h$ segments and integrates the output of these individual heads.
\begin{equation}
\text{MultiHead}(\mathbf{Q}, \mathbf{K}, \mathbf{V}) = \text{Concat}(\text{head}_1, \dots, \text{head}_h)\mathbf{W}_O,
\end{equation}
where $\text{head}_j = \text{Attention}(\mathbf{Q}_j, \mathbf{K}_j, \mathbf{V}_j)$ and \( \mathbf{W}_O \) is the learned projection matrix.

After the Transformer encoder is deployed on temporal input data, the output representation $\mathbf{H}^L$ captures intricate temporal inter-dependencies across various timestamps. Next we design a decoder to elevate the model's proficiency in processing complex sequence data for action prediction. The decoder architecture incorporates a multi-head cross-attention module. It is operated by dynamically focusing on correlated segments of the historical trajectory in relation to the current state $\mathbf{s}_t$. The structural and functional dynamics of the cross-attention module are similar to the previous self-attention module. The decoder output representation matrix $Z^{\prime}_i$ of trajectory $\tau_i$ is computed as follows.

\begin{equation}
\mathbf{Z^{\prime}_i} = \text{softmax}\left(\frac{\mathbf{Q}_i\mathbf{K}_i^T}{\sqrt{d_k}}\right)\mathbf{V}_i
=\text{softmax}\left(\frac{(\mathbf{S}_i\mathbf{W}_q^{\prime})(\mathbf{Z}^L_i\mathbf{W}_k^{\prime})^T}{\sqrt{d_k}}\right)(\mathbf{Z}^L_i\mathbf{W}_v^{\prime}).
\end{equation}

In the above equation, $\mathbf{Z_i}^L$ denotes encoding historical information of trajectory $\tau_i$, $\mathbf{Z_i}^L = \{\mathbf{z}_1, ..., \mathbf{z}_T\}$, and $\mathbf{S}_i$ denotes the broadcasting matrix of the current state $s_t$. The output of multi-head cross attention module is also obtained by concatenating the outputs from all heads and projecting them through a linear layer.

In the unique scenario of carbon neutral optimization task, we face a continuous action space that encompasses multiple inter-related action types. Unlike traditional solutions to employ the sigmoid function for categorical prediction, our objective is to allocate a distinct value to each type of action. Hence we create an action vector while simultaneously expanding the exploration space for these actions. Inspired by previous reinforcement learning studies~\cite{chua2018deep, agostini2010reinforcement}, we make the assumption that the actions adhere to a multivariate distribution. Based on this distribution learned by the output layer, we sample the recommended action $a$ at the current timestamp, as shown in the following equation.
\begin{equation}
    \pi_\theta (a|h, s)  = p(a; \mu, \Sigma)
\end{equation}
where $\mu \in R^K$ and $\Sigma \in R^{K \times K}$ represent the mean vector and covariance matrix of the actions respectively. They are the output of the dense layer with the input $Z^{\prime}$.

\subsection{Environment Simulator}
The decision-making process often requires a precise prediction of the state evolutions following conducted actions in the trajectory. In pursuit of modeling this predictive framework, we employ the Variational Auto-Encoder (VAE)~\cite{kingma2013auto}, a deep generative model that captures potential future states following the actions. For a given state \( s_t \) and a designated action \( a_t \), the encoder of the VAE maps the state-action pairing into a latent space as:
\begin{equation}
q_\phi(z|s_t, a_t) = \mathcal{N}(z; \mu(s_t, a_t), \sigma^2(s_t, a_t))
\end{equation}
where $\mu$ and $\sigma$ are mean and variance of $z$.

The core objective during training is to approximate the true posterior distribution. It aims to minimize the discrepancy between predicted future states and observed outcomes, formulated as \( p_\theta(s_{t+1}|s_t, a_t, z) \). The VAE integrates a regularization term to prevent overfitting and ensure a smoother latent space:
\begin{equation}
\mathcal{L} = \mathbb{E}_{q_\phi}[\log p_\theta(s_{t+1}|s_t, a_t, z)] - \beta \text{KL}(q_\phi(z|s_t, a_t) || \mathcal{N}(0, I))
\end{equation}
where $\beta$ is a hyper-parameter to balance the two terms in the loss function. The Environment Simulator is trained prior to the imitation learning framework. Once it achieves a predefined performance criterion, its parameters are fixed to ensure consistent interactions during subsequent learning stages. With this simulation framework, it becomes feasible to predict the consequences of certain actions, thereby informing the decision-making process and paving the way for crafting optimal operational sequences \cite{ha2018recurrent, gomez2018automatic}.

\subsection{Policy Optimization}
The major training objective of PAIL is to make the recommended actions close to the ones from trajectories of high SGD. Meanwhile, the system should make the value of each action as large as possible to improve the generalization ability. To this end, we design a dual policy optimization strategy consisting of two components, namely the discriminator and performance estimator.

\subsubsection{Discriminator of PAIL} 
PAIL draws inspiration from the foundational principles of Generative Adversarial Networks (GAN) ~\cite{goodfellow2020generative} to train policies in reinforcement learning. Just like GAN, PAIL uses a discriminator to distinguish between the trajectories with high SGD and those ones generated by the model. By trying to obfuscate the discriminator, the policy generator can effectively learn to imitate good trajectories. 

PAIL integrates the historical information together with current state to generate the recommended actions. The historical vector is got by applying average pooling on the original vectors, denoted as h.
PAIL leverages the $\omega$-parameterized multiple-layer perception (MLP) $D_\omega(h, s, a)$. This function estimates the likelihood of a given history-state-action tuple $(h, s, a)$ originating from a trajectory of high SGD.
The discriminator is addressing a binary classification task, and both the policy generator and the discriminator engage in a Min-Max game predicated on the cross-entropy loss.

\noindent The objective of discriminator is:
\begin{equation}
\max_\omega \mathbb{E}_{\pi_{\text{E}}}\left[\log D_\omega(h, s, a)\right] + \mathbb{E}_{\pi_{\theta}}\left[\log(1 - D_\omega(h, s, a))\right]
\end{equation}
\noindent The objective of policy generator is:
\begin{equation}
L_{IL} = \min_\theta \mathbb{E}_{\pi_{\theta}}\left[\log (1 - D_\omega(h, s, a))\right]
\end{equation}
In real world applications, the trajectories with high SGD are typically much rare than the ones with middle or low SGDs. The limited training data pose a big challenge to proposed imitation learning framework. To address this issue, we propose a performance-oriented training guidance mechanism for the policy generator. This mechanism aims to maximize the cumulative SGD value of each state-action pair, enhancing the overall performance of generated trajectories. The key is to derive a reward signal for each discrete timestamp and interpret the discriminator output as an inverse measure of the policy performance. Formally, the reward signal for a history-state-action tuple $(h, s, a)$ is quantified as $R(h, s, a) = -\log D_\omega(h, s, a)$. This formulation establishes an intriguing relationship: When the discriminator assigns a value close to 0, indicating significant deviation from existing trajectories of high SGD, the corresponding reward is a large negative value. This penalizes the policy generator for low-SGD actions, guiding it towards high-SGD recommendations. Subsequently, we develop a performance estimator to assess the value of each history-state-action tuple.

\subsubsection{Performance Estimator}
In real industrial systems, quantifying the immediate SGD credit of each single action in a long trajectory is non-trivial. In most scenarios, the historical trajectories only have a final SGD value as the overall performance. This challenge of attributing SGD credit to distinct actions has led us to exploit Temporal Difference (TD) learning for a deep $Q$-network~\cite{mnih2013playing}. The network aims to estimate the utility of any given state-action pair. Specifically, we propose a refined self-attention network $F$ to capture historical inter-dependence across time and assess the SGD credits for all the actions in the trajectory.
The network architecture is similar to the encoder of policy generator. It is pre-trained by minimizing the square loss to existing trajectories with high SGD. For trajectory $\tau$ with length of $N$, the overall SGD is calculated as:
\begin{equation}
    V(\tau) = F (s_1, a_1, \cdots, s_N, a_N)
\end{equation}

Hence the immediate reward at timestamp $t$ is obtained by discriminator as follows.
\begin{equation}
r_t = -\mathbb{E}_{(h,s,a) \sim \pi_\theta} \log(D_\omega(h_t, s_t, a_t))
\end{equation}

In order to evaluate the SGD credit of a specific state-action pair, we initialize a $Q$-network, denoted by $Q(s,a|\theta^Q)$, with arbitrary parameters. Meanwhile, we define the target value network as $Q^{\prime}(s,a|\theta^{Q^{\prime}})$, which helps to stabilize the learning process by providing a fixed baseline against which the predictions of the $Q$-network can be compared, reducing the risk of feedback loops and divergence during training~\cite{mnih2015human}. The Temporal Difference (TD) error, crucial for updating the Q network, is calculated as follows:
\begin{align}
\delta_t &= r_t + \gamma Q'(s_{t+1}, a_{t+1}|\theta^{Q'}) - Q(s_t, a_t|\theta^Q)\\
\delta_T &= r_T - Q(s_T, a_T|\theta^Q) = V(\tau) - Q(s_T, a_T|\theta^Q)
\end{align}
where $\gamma$ represents the discount factor to weight the importance of future rewards relative to immediate ones, effectively capturing the present value of future SGD credits, and $Q'(s_{t+1}, \pi_{\theta}(s_{t+1})|\theta^{Q'})$ denotes the SGD credit of next state-action pair estimated by the target network. To encourage exploration and prevent premature convergence to sub-optimal policies, PAIL introduces Gaussian noise to the deterministic action output of the policy network and uses them as the input to the target network $Q^{\prime}$. In this way, PAIL facilitates strategic exploration of the action space to optimize $Q$-value of subsequent state-action pairs.

The goal of TD learning is to minimize the temporal difference (TD) error. Accordingly, the loss function for TD learning module is formulated as:
\begin{equation}
\mathcal{L}(\theta^{Q}) = \mathbb{E}_{(s,a) \sim \pi_\theta} \left[ \delta_t^2 \right],
\end{equation}
where $\delta_t$ represents the TD error at time $t$. The parameter updating steps of both $Q$ and $Q^{\prime}$ networks are synchronized with the imitation learning module. This learning process involves multiple iterations of updates in a single epoch, delineated as follows:
\begin{align}
\theta^Q &\leftarrow \theta^Q - \eta \nabla_{\theta^Q} \mathcal{L}(\theta^Q), \\
\theta^{Q'} &\leftarrow \epsilon \theta^Q + (1 - \epsilon) \theta^{Q'},
\end{align}
where $\eta$ denotes the learning rate for gradient descent in the module, and $Q'$ network is softly updated by copying the parameters from the $Q$ network. $\epsilon$ is set as a small constant (e.g., 0.01) to ensure gradual updates.

In essence, the goal is to optimize a neural architecture and infer the SGD credits of different state-action pairs. The guide is on the maximization of the $Q$-value for recommended actions, as shown in the following equation.
\begin{equation}
L_{value} = -E_{a \sim \pi_\theta(\cdot|s)} [Q(s, a)]
\end{equation}
Thus, the loss function for the policy generator is:
\begin{equation}
    L = \lambda L_{IL} + (1 - \lambda) L_{value} + \beta(t) H(\pi)
\end{equation}
where $\lambda$ is a hyper-parameter to moderate the relative importance of the two objectives, and $H(\pi)$ is the entropy of learned policy. In PAIL framework, the entropy regularization term is dynamically adjusted using a decay function for the time-dependent coefficient $\beta(t)$. PAIL uses an exponential decay function $\beta(t) = \beta_0 \cdot e^{-kt}$, where $\beta_0$ is the initial value, $k$ is the decay rate, and $t$ represents the epoch. This exponential decay allows for aggressive exploration in the initial phases of training and progressively shifts the focus towards exploitation by reducing the influence of the entropy over time. As a result, the policy generator derives its learning signal by back-propagating from both the discriminator and the performance estimator. This dual influence ensures that the generator not only emulates the trajectories of high SGD but also ensures each action is optimized for highest credit.

\begin{figure*}[ht]
    \centering
    \includegraphics[width=0.95\linewidth]{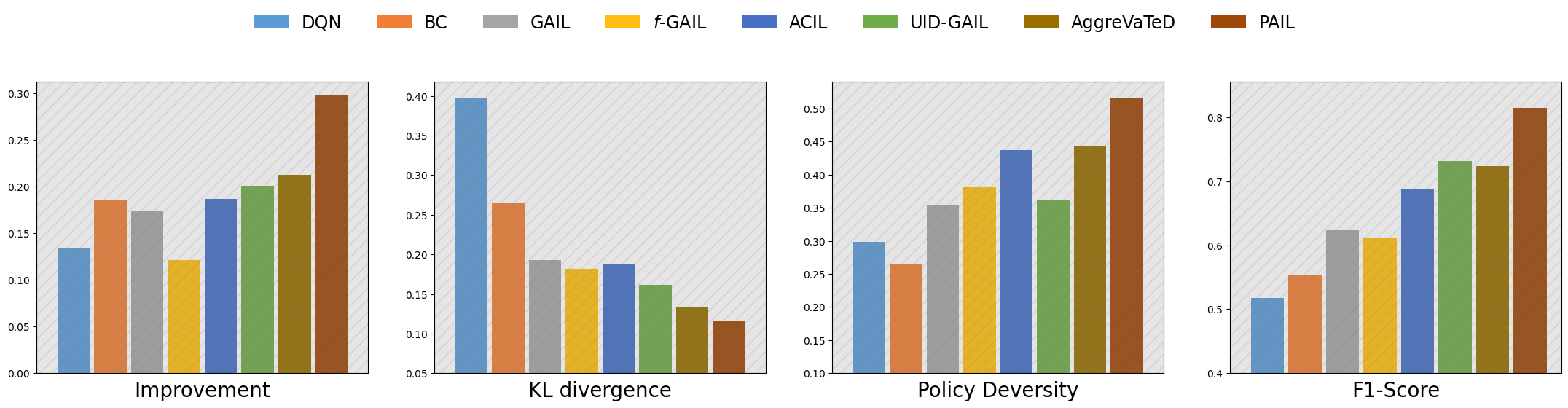}
    \caption{The Overall performance of baselines and PAIL for the oil production dataset.}
    \label{op}
\end{figure*}

\begin{figure*}[ht]
    \centering
    \includegraphics[width=0.95\linewidth]{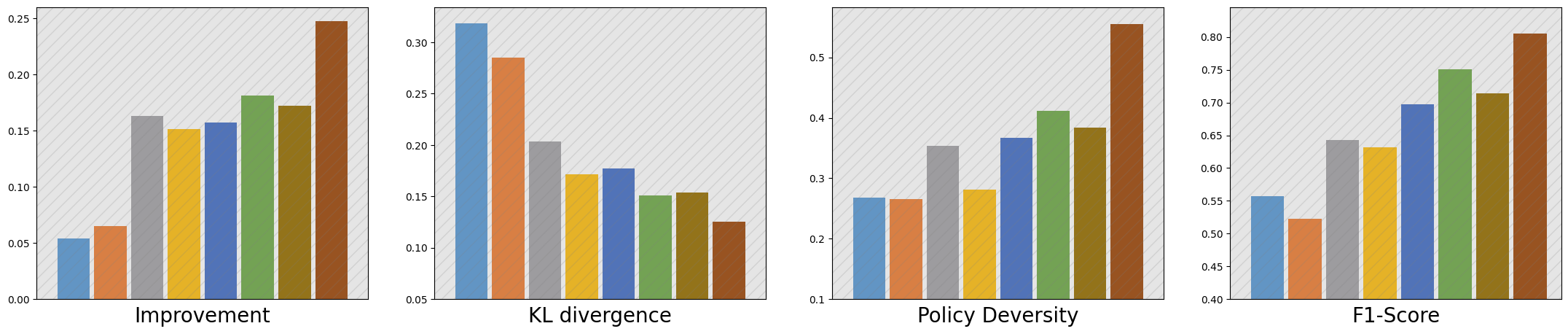}
    \caption{The Overall performance of baselines and PAIL for the supply chain dataset.}
    \label{ot}
\end{figure*}

\section{Experiments}
In this section, we introduce the details of our experiments conducted on two industry datasets of carbon neutral optimization.

\vspace{-2mm}
\subsection{Experimental Settings}
\subsubsection{Dataset}

In this study, we conducted an empirical analysis using two distinct datasets. The first dataset was collected from the gas production industry. The dataset consists of 1,100 production processes, each process is with 30 steps (i.e., time window to carry out actions). The sensors record key parameters like system temperature and air pressure. From a broader set of metrics, we selectively filtered the dataset to include the most informative 51 parameters, which collectively define the state space. Additionally, the dataset delineates the action space available at each decision point, encompassing five distinct types of actions. The final Sustainable Development Goal (SDG) is measured by considering both economic efficiency and carbon credits. Each data sample has only one SDG at the end to evaluate the overall process.

The second dataset was sourced from a supply chain system, comprising 500 transaction sequences. Each sequence is divided into 30 time windows. The data captures purchase actions by 5 factories aimed at meeting buyer demands, including details on the acquired volume, average cost, and $CO_2$ emissions for each transaction. Accordingly, the buying behavior of each factory is identified as the primary action within this multi-dimensional space. To complement this, 15 additional critical indicators have been selected and incorporated to construct the state space. Consequently, an SDG indicator is also included to assess the entire transaction sequence, taking into account both the transaction amount and carbon credits.

For both datasets, we selected the sequences of top $10\%$ as high-performance trajectories and used all of them to train the model. For the remainder data, we partitioned them at a 4:1 training/test ratio for the experiment.

\subsubsection{Metrics}
To evaluate the proposed model, we used a variety of metrics. Firstly, we employed a pre-trained trajectory value estimator for predicting the \textbf{improvement} ratio of SDG compared to the benchmark, utilizing off-policy evaluation to gauge the value of learned policy without real interactions. Specifically, we employed the Doubly Robust Off-policy Value Evaluation for an unbiased policy value estimation, merging direct method estimation with importance sampling for greater accuracy. Consistent with previous imitation learning studies, we measured the discrepancy between generated and optimal policies using \textbf{KL divergence}, and assessed \textbf{policy diversity} through the complement cosine similarity of generated action sequences. Moreover, we used the \textbf{F1-score} to assess improvements in terms of individual trajectories by setting a threshold that classified the top 50\% as positive samples and the remainder as negative before updating the policy, in order to track transitions from negative to positive statuses while monitoring the retention of positive samples.

\subsubsection{Baseline Methods}
We compared the performance of PAIL with several state-of-the-art methods. In the following list, we briefly introduce these baselines.
\begin{itemize}
    \item \textbf{DQN}: DQN~\cite{mnih2015human} trains policy via deep Q-learning by designing a sparse reward function.
    \item \textbf{Behavior Cloning}: BC~\cite{pomerleau1989alvinn} cuts the trajectories into state action tuples and learns a policy from the demonstrations by supervised learning.
    \item \textbf{GAIL}: GAIL~\cite{ho2016generative} generates a policy by mimicking the expert trajectories via the reward signals of the discriminator.
    \item \textbf{$f$-GAIL}: $f$-GAIL~\cite{zhang2020f} dynamically learns the optimal $f$-divergence for more effective policy learning.
    \item \textbf{ACIL}: ACIL~\cite{wang2020adversarial} utilizes both positive and negative demonstration through adversarial and cooperative discrimination of trajectories.
   \item \textbf{AggreVaTeD}: AggreVaTeD~\cite{sun2017deeply} leverages historical trajectories for imitation learning, integrating differentiable function approximators to facilitate sequential decision making.
    \item \textbf{UID-GAIL}: UID-GAIL~\cite{wang2023unlabeled} learns from imperfect demonstrations by treating them as unlabeled data and applying a positive-unlabeled learning approach.
\end{itemize}
To ensure fair comparisons, we standardized the starting step of action update at $t_s = 9$ for both datasets, and uniformed the embedding dimensions for historical data, states, and actions to 64. For enhancing the reliability of our findings, f each baseline and our proposed method, we conducted experiments using three distinct initial seeds and employed five-fold cross-validation. The reported results represent the average across these configurations, ensuring a 95\% confidence level.
\vspace{-1mm}

\subsection{Performance Analysis}
Figure~\ref{op} and~\ref{ot} show the performance of PAIL and baselines evaluated by the four metrics on both datasets. From the results, we have the following observations: (1) DQN demonstrated limited effectiveness in environments with sparse reward signals, highlighting a critical area for learning policy with developing imitation learning based methods. (2) Behavioral Cloning (BC) suffered across all metrics due to poor generalization caused by compounding errors and heavy reliance on high-quality expert data, which cannot be available in our problem definition. (3) While classical GAIL and its variants showed reasonable performance on metrics like KL divergence, their overall effectiveness was compromised in environments requiring nuanced action choices, likely due to poor generalization, which suggests a critical need for strategies that extend beyond mere replication of observed behaviors in the context of carbon neutrality. (4) Among previous imitation learning methods, AggreVaTeD distinguished itself by utilizing historical information to enhance learning outcomes, and UID-GAIL also showed promising results by adaptively utilizing imperfect demonstrations to learn policy, suggesting the value of historical context and auxiliary guidance in industrial operation optimization. (5) Finally, PAIL outperformed existing benchmarks across all metrics, demonstrating superior aggregate and individual performance, as well as enhanced generalization capabilities to unseen sequences.
\vspace{-1mm}
\subsection{Ablation Study}
\begin{figure}[ht]
    \centering
    \begin{subfigure}{0.42\columnwidth}
    \includegraphics[width=\columnwidth, height=0.75\columnwidth]{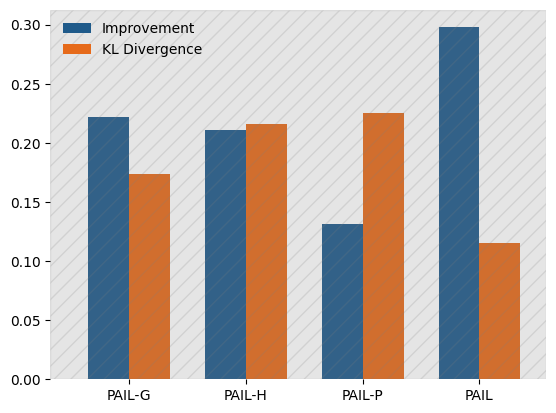}
    \caption{Oil Production}
    \end{subfigure}
    \begin{subfigure}{0.42\columnwidth}
        \includegraphics[width=\columnwidth, height=0.75\columnwidth]{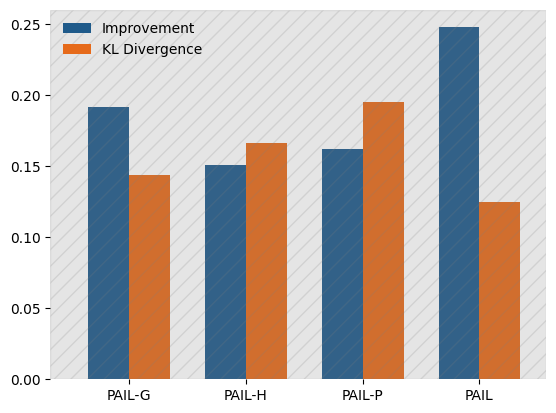}
        \caption{supply chain}
		\end{subfigure}
	\caption{Performance comparison of PAIL and variants.}
	\label{fig:ablation}
\end{figure}

To evaluate the contributions of various components in the PAIL framework, we developed several variants of PAIL. Their descriptions and modifications are listed as follows:
\begin{itemize}
\item \textbf{PAIL-G}: In this variant, we removed the probabilistic sampling via the Gaussian mixture model and took the deterministic output to generate the policy.
\item \textbf{PAIL-H}: This variant simplified the input to the discriminator by utilizing only the state-action pair, instead of using history-state-action tuples.
\item \textbf{PAIL-P}: This variant omitted the Q-learning based performance estimator module.
\end{itemize}
As depicted in Figure~\ref{fig:ablation}, a comparative evaluation between PAIL and PAIL-G underscored the importance of probabilistic sampling from a distribution to prevent the model from prematurely falling into suboptimality. Additionally, PAIL-H exhibited notably inferior performance relative to integrating history in evaluating the discrepancy, specifically in oil production data, which further emphasized the advantages of taking historical information into account. When compared with PAIL, the worse performance of PAIL-P validated the efficacy of the proposed dual policy refinement mechanism, which integrates a Q-learning module for targeted modeling on value promotion of each distinct action.

\subsection{Parameters Tuning}
\subsubsection{Impact of $t_s$} 
\begin{figure}[t]
    \centering
    \begin{subfigure}{0.41\columnwidth}
    \includegraphics[width=\columnwidth]{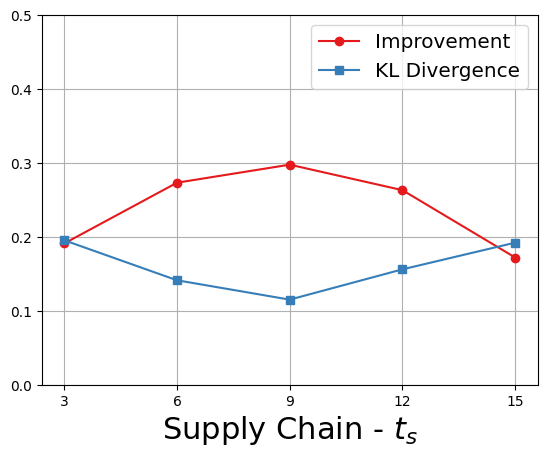}
    \end{subfigure}
    \begin{subfigure}{0.41\columnwidth}
        \includegraphics[width=\columnwidth]{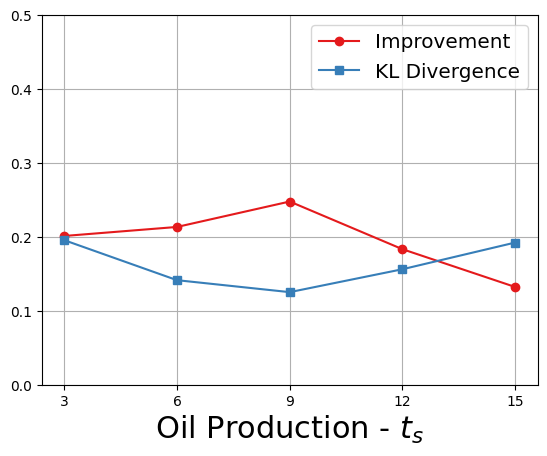}
		\end{subfigure}
	\caption{Performance of PAIL under different $t_s$.}
	\label{fig:sp}
\end{figure}

In order to assess the robustness of PAIL, here we evaluated our model under different starting point $t_s$. As Figure~\ref{fig:sp} shows,  PAIL exhibited relatively poor performance on both datasets under conditions where the threshold $t_s$ was either large or small. This trend could be attributed to inadequate learning from limited historical data and sequence features at lower $t_s$ values, and restricted update capability at higher $t_s$ values.

\begin{figure}[t]
    \centering
    \begin{subfigure}{0.41\columnwidth}
    \includegraphics[width=\columnwidth]{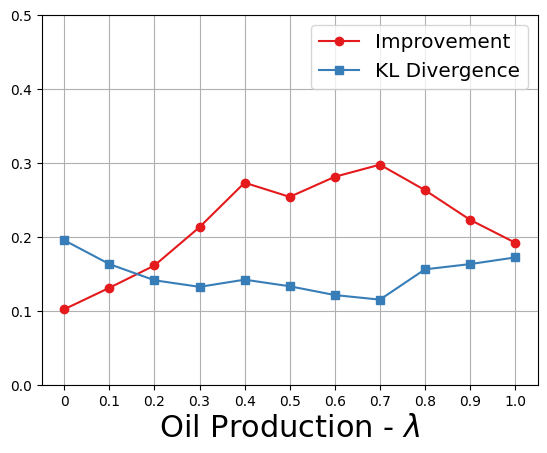}
    \end{subfigure}
    \begin{subfigure}{0.41\columnwidth}
        \includegraphics[width=\columnwidth]{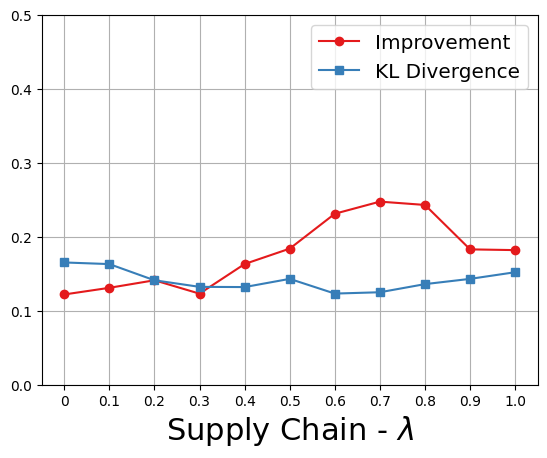}
		\end{subfigure}
	\caption{Performance of PAIL under different $\lambda$.}
	\label{fig:lambda}
\end{figure}

\begin{figure}[t]
    \centering
    \begin{subfigure}{0.41\columnwidth}
    \includegraphics[width=\columnwidth]{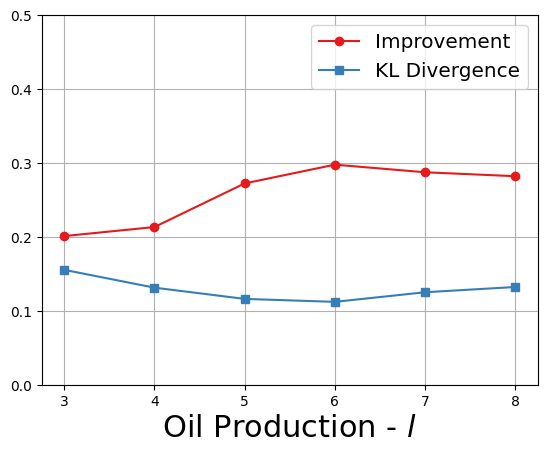}
    \end{subfigure}
    \begin{subfigure}{0.41\columnwidth}
        \includegraphics[width=\columnwidth]{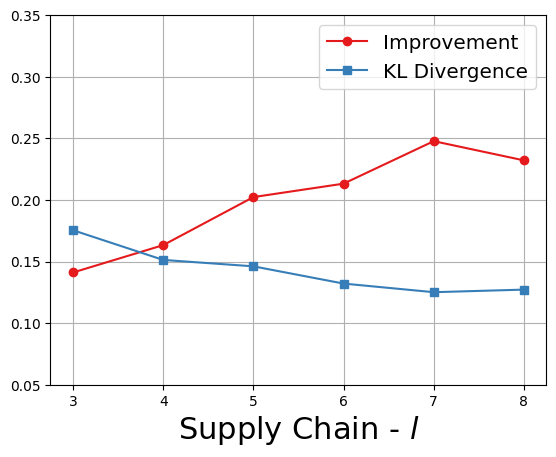}
		\end{subfigure}
	\caption{Performance of PAIL under different $l$.}
	\label{fig:l}
\end{figure}

\subsubsection{Impact of $\lambda$} Subsequently, we examined the influence of parameters $\lambda$ to evaluate the importance of two objectives for policy refinement. As Figure~\ref{fig:lambda} shows, the performance significantly declined when the lambda parameter was either too large or too small, demonstrating the effectiveness of our designed discriminator module and performance estimator module in aiding policy generation. When $\lambda$ was too small, despite the TD learning module utilizing information from the discriminator, relying solely on the learned Q-function proved to be insufficiently robust. Furthermore, the performance degradation observed with a large lambda further indicated that an approach based solely on imitation learning principles leaded to generalization issues.

\subsubsection{Impact of $l$}: Figure~\ref{fig:l} depicts the investigate on the impact of the lookback size $l$ on model performance when considering the historical information. Initially, as the lookback length $l$ increased, the performance of PAIL improved, highlighting the importance of historical information for determining the current optimal action to achieve carbon neutral. However, as $l$ continued to increase, the model performance tended to stabilize and may even slightly decline. We speculated this may be due to the encoding of information irrelevant to the current state, where an overload of information leaded to fluctuations in the model's predictive performance.

\vspace{-1mm}

\subsection{Case Study}
\begin{figure}[t]
    \centering
    \includegraphics[width=0.85\columnwidth]{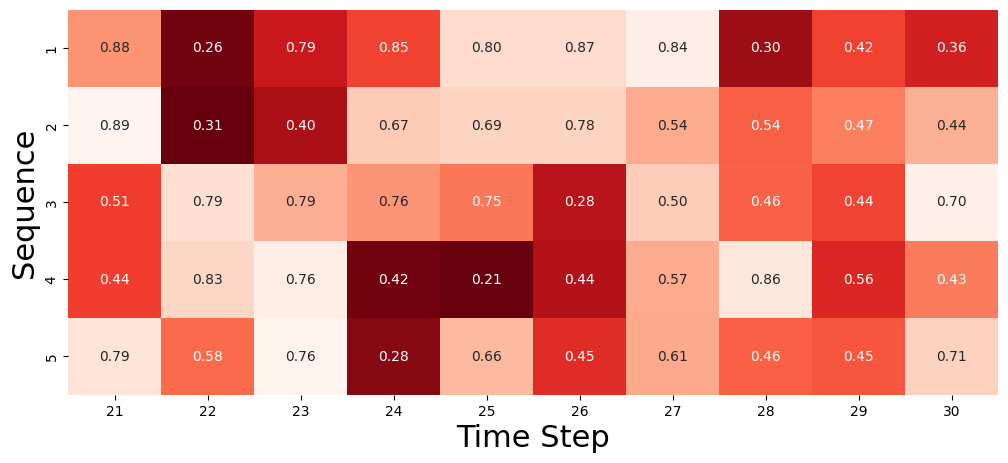}
    \vspace{-3mm}
    \caption{Time step importance and action vector similarity.}
    \vspace{-5mm}
    \label{fig:case}
\end{figure}

In this section, we present a case study where we identified several trajectories with significant SGD improvements. As Figure~\ref{fig:case} shows, we created a heatmap based on the importance of distinct time steps to the final SGD. A segment of this heatmap is highlighted, where darker colors indicate greater importance of a given time step. Subsequently, we quantified the similarity between the updated action vector at each step and the original action vector. Our observation revealed that areas with darker colors corresponded to lower similarity scores, indicating more substantial changes in actions. This pattern further substantiated that our method could effectively alter sequences in a manner that enhances the corresponding SGD performance, highlighting its potential to optimize learning processes through strategic modifications in action sequences.

\section{Related Work}
\subsection{KDD for Social Good}
Recent advances in machine learning and data mining have significantly contributed to various social good domains, including healthcare~\cite{komorowski2018artificial, wang2018supervised,zheng2021drug}, smart cities~\cite{yuan2020spatio, zhang2020semi}, and financial services~\cite{fang2023learning, liu2021intention, yu2023cognitive, yu2022collaborative}. Key studies~\cite{vinuesa2020role, hager2019artificial, shi2020artificial} have underscored AI's critical role in enhancing decision-making across socioeconomic spheres, particularly at the nexus of economic efficiency and environmental sustainability. Research, notably~\cite{li2015energy}, has leveraged advanced machine learning to link economic growth with energy consumption, aligning with the United Nations Sustainable Development Goals (SDGs) and highlighting the contribution of AI to sustainable development. A broad spectrum of AI-driven research focuses on optimizing energy consumption through intelligent devices~\cite{fiducioso2019safe}, user behavior modeling~\cite{li2015energy, ye2019identifying, ye2022mane, ye2024university}, and smart home technologies~\cite{truong2013forecasting}. Extending beyond energy, this body of work explores water conservation and system optimization, alongside enhancements in electric vehicle fuel efficiency~\cite{vogel2012improving, wu2018efficiently} and strategic charging station placement~\cite{valogianni2015multiagent, zhang2022rlcharge}. Our study applies imitation learning to improve industrial processes towards carbon neutrality, providing novel AI insights for social good in line with the SDGs, showcasing the dual benefits of energy conservation and economic advantage for a sustainable future.

\vspace{-2mm}
\subsection{Imitation Learning}
Recent advances in deep reinforcement learning (DRL) have significantly propelled efforts to optimize sequential decision-making across various fields, including autonomous vehicles~\cite{kiran2021deep, shalev2016safe}, treatment recommendations~\cite{komorowski2018artificial, prasad2017reinforcement,zheng2023interaction}, and economics~\cite{jiang2017deep, yang2020deep}, extending to industry operation system optimization~\cite{liu2020parallel, han2020enabling}. The primary aim of DRL research has been to develop a learning policy driven by specific rewards. Yet, crafting such reward signals has often been challenging due to the necessity for complex calculations, especially in long-term optimization scenarios where immediate indicators are insufficient.

Thus, Imitation Learning (IL) has emerged as a vital subset of reinforcement learning (RL), emphasizing learning from expert policies to derive reward signals. This approach has been applied in diverse areas from video games~\cite{fei2020triple} to autonomous driving~\cite{bhattacharyya2022modeling, choi2021trajgail} and robotics~\cite{sutton2018reinforcement, chen2020learning}. IL includes Behavioral Cloning (BC), which maps observations to actions through supervised learning~\cite{pomerleau1989alvinn}, and Inverse Reinforcement Learning (IRL), which infers experts' implicit reward functions to grasp their decision-making process~\cite{ng2000algorithms, abbeel2004apprenticeship}. Although BC has suffered from compounding errors~\cite{ross2011reduction} and IRL from computational complexities, Generative Adversarial Imitation Learning (GAIL) has combined their strengths to match the learner's state-action distributions with those of the expert, mitigating their individual drawbacks~\cite{ho2016generative,zheng2023generative}. Subsequent GAIL variants, such as AGAIL~\cite{sun2019adversarial}, which learns from incomplete demonstrations using partial actions, and Triple-GAIL~\cite{fei2020triple}, which introduces skill selection and multi-modal imitation, have shown enhanced adaptability in complex settings. Nonetheless, IL's dependence on high-quality demonstrations and its sensitivity to distributional shifts, where policies may face unseen states, presents notable challenges. These limitations have highlighted concerns over IL's generalization capabilities, underscoring the difficulty in ensuring optimal policy generation across all industrial scenarios due to ambiguous distinctions between positive and negative samples in our task context.

\section{Conclusion}
In this study, we address the critical challenge of achieving carbon neutrality in industrial operations, presenting a novel Performance-based Adversarial Imitation Learning (PAIL) engine. By leveraging Imitation Learning techniques, we design a Transformer based policy generator to produce operational policies without relying on pre-defined action rewards. Then, through dual reward mechanism refinement with utilizing an adapted discriminator and a Q-learning based performance estimator, PAIL updates policies aligned with the optimal sustainable development goals (SDG). Extensive experiments across various real-world scenarios demonstrate the effectiveness of PAIL compared to other state-of-the-art methods, offering both enhanced performance and interpretability in achieving carbon neutrality.
\section*{Acknowledgments}
This research was done during an internship at NEC labs, and partially supported by the National Science Foundation (NSF) via the grant number IIS-2006387 and IIS-2040799.


\bibliographystyle{ACM-Reference-Format}
\bibliography{acmart}



\end{document}